\documentclass[cameraready]{Interspeech}

\title{Who Spoke What When? Evaluating Spoken Language Models for Conversational ASR with Semantic and Overlap-Aware Metrics}

\author[affiliation={1}, orcid=0000-0002-1130-5059]{Naohiro}{Tawara}
\author[affiliation={2}, orcid=0000-0002-5358-1844]{Samuele}{Cornell}
\author[affiliation={2,3}, orcid=0009-0000-4958-202X]{Alexander}{Polok}
\author[affiliation={1}, orcid=0000-0002-5175-7834]{Marc}{Delcroix}
\author[affiliation={3}, orcid=0000-0002-4951-5908]{\\Lukáš}{Burget}
\author[affiliation={2}, orcid=0000-0000-0000-0000]{Shinji}{Watanabe}

\address{
    $^1$ NTT, Inc., Japan, 
    $^2$ CMU, USA, 
    $^3$ BUT, Czechia
}

\email{naohiro.tawara@ntt.com}

\keywords{conversational speech recognition, speaker diarization, speech language models, semantic evaluation metrics}

\usepackage{comment}
\usepackage{graphicx}
\usepackage[table]{xcolor}
\usepackage{bm}
\usepackage{cite}
\usepackage{amsmath,amssymb,amsfonts,mathrsfs}
\usepackage{bbm}
\usepackage{amsxtra}
\usepackage{threeparttable}
\usepackage{algorithm,algorithmic}
\usepackage{etoolbox,siunitx}
\usepackage{multirow, booktabs}
\usepackage{balance}
\usepackage{textcomp}
\usepackage{xcolor}
\usepackage{url}
\usepackage{pifont}
\usepackage{threeparttable}
\usepackage{hyperref}
\usepackage{makecell}
\usepackage{setspace}
\usepackage{colortbl}
\usepackage[normalem]{ulem}
\usepackage{subcaption}
\usepackage{enumitem}
\newcolumntype{H}{>{\setbox0=\hbox\bgroup}c<{\egroup}@{}}
\usepackage{siunitx}
\newcolumntype{P}{S[round-mode=places, round-precision=1, table-format=1.2]}
\newcommand{\todo}[1]{\textcolor{red}{\textbf{TODO:} #1}}
\newcommand\defm[2]{\expandafter\newcommand{#1}{\ensuremath{#2}}} 
\defm\R{R}
\defm\hyp{H}

\robustify\bfseries

\begin{document}

\maketitle

\begin{abstract}
Conversational automatic speech recognition remains challenging due to overlapping speech, far-field noise, and varying speaker counts. While recent LLM-based systems perform well on single-speaker benchmarks, their robustness in multi-speaker settings is unclear. We systematically compare LLM-based and modular pipeline approaches along four axes: overlap robustness, semantic fidelity, speaker count, and single- versus multi-channel input. To capture meaning-altering errors that conventional metrics miss, we introduce tcpSemER, which extends tcpWER by replacing Levenshtein distance with embedding-based semantic similarity. We further decompose tcpWER into overlapping and non-overlapping components for finer-grained analysis. Experiments across three datasets show that LLM-based systems are competitive in two-speaker settings but degrade as speaker count and overlap increase, whereas modular pipelines remain more robust.
\end{abstract}

\section{Introduction}
\label{sec:intro}

Despite the remarkable progress in automatic speech recognition (ASR)~\cite{li2022recent, prabhavalkar2023end, whisper}, multi-talker conversational speech remains one of the central open problems in speech processing~\cite{cornell2025recent, abramovski2025summary}.
Conversational ASR (CASR) aims to determine \emph{who spoke what and when} in multi-party recordings by integrating speaker diarization with ASR~\cite{haeb2020far, cornell2025recent, abramovski2025summary, he2025survey}.
Modular pipelines~\cite{niu2024ustc, wangustc, mitrofanov2024stcon, ye2023iacas} (e.g., cascade of diarization, separation, and ASR) remain the dominant approach and have demonstrated substantial progress in recent evaluation campaigns, such as CHiME-7/8 Distant ASR (DASR)~\cite{cornell2025recent} and NOTSOFAR-1~\cite{abramovski2025summary, vinnikov24_interspeech}.

Recently, LLM-based approaches have emerged as an alternative. They can be divided into two categories.
\emph{Task-specific} LLM-based systems~\cite{peng2026vibevoice, yu2026moss, huo2026tagspeech, voxtral2026} fine-tune an LLM backbone for end-to-end (E2E) long-form multi-speaker transcription, becoming thus de facto E2E transcription systems~\cite{chang2025module}. These systems benefit from a more powerful backbone and the advantage of long-form processing compared to modular ones~\cite{cornell2024one, niu2025dcf, hirano24_chime, huang24b_chime, hu24_chime, polok24_chime, broughton24_chime,kalda24_chime, kanda2022transcribe, zheng2025dncasr, park2025sortformer, nguyen2025msa}.
However, they typically lose general LLM capabilities such as summarization or question-answering (QA)~\cite{yu2026moss, peng2026vibevoice, voxtral2026}. 
\emph{General-purpose multimodal} LLMs such as Gemini~\cite{comanici2025gemini, gemini3flash2025} can be prompted to generate speaker-attributed transcripts as one of many capabilities.
Such systems offer versatility through zero-shot generalization and the potential for improved performance by leveraging diverse training data and tasks.

It remains unclear how robustly each category handles the specific challenges of CASR compared to state-of-the-art (SotA) modular systems, particularly in the presence of overlapping speech, distant microphone recordings, and long-form audio.
To address this gap, we present a systematic comparison of a general-purpose multimodal LLM (Gemini~3.0 Flash~\cite{gemini3flash2025}),  task-specific LLM-based systems (VibeVoice~\cite{peng2026vibevoice}, Voxtral Mini Transcribe v2~\cite{voxtral2026}), and SOTA modular pipelines in both single-~\cite{polok2025dicow,han2025diarizen} and multi-channel~\cite{kamo2025ntt_chime8} configurations across three conversational datasets, introducing new evaluation metrics tailored to the task. 

A meaningful comparison, requires evaluation metrics that extend beyond standard word error rate (WER) formulations for CASR.
Concatenated minimum-permutation WER (cpWER)~\cite{watanabe2020chime} and its time-constrained variant (tcpWER)~\cite{von2023meeteval, vonneumann2025wer} treat all word errors equally, irrespective of their semantic impact and are highly sensitive to the text normalization scheme adopted. 
This is a significant limitation in conversational speech, where fillers, repetitions, and disfluencies are prevalent and is even more problematic when evaluating LLM-based systems, which often generate ``standardized'', ``fluent'' output that may diverge from verbatim transcription even if preserves meaning. Previous embedding-based metrics, such as SemDist~\cite{kim2021semantic} and SeMaScore~\cite{sasindran2024semascore}, attempt to address these shortcomings of WER. However, they are fundamentally restricted to single-speaker, utterance-level evaluation.
Specifically, they assume fixed reference–hypothesis alignment at the utterance level and do not account for permutation invariance or temporal constraints, which are essential in multi-speaker long-form CASR.
To overcome these limitations, we propose an extension of tcpWER, called time-constrained minimum permutation semantic error rate (tcpSemER), which measures embedding-based semantic similarity between temporally aligned speech segments. 
Moreover, as standard metrics do not differentiate between errors in overlapping and non-overlapping regions, we also introduce an overlap-aware decomposition of tcpWER and cpWER. 

We open-source the code for all proposed metrics and release a public leaderboard.\footnote{URL removed for double-blind review.}
Through this analysis, we clarify the limitations and potential of the different CASR frameworks. 


\section{Experimental Protocol}
\label{sec:protocol}


\subsection{Systems Under Evaluation}
\begin{itemize}
\item \textbf{Modular Systems:}
For single-channel, we use the \emph{DiCoW} system, which combines the DiariZen diarization front-end~\cite{han2025diarizen}, based on E2E neural diarization with vector clustering (EEND-VC)~\cite{eend-vc, Bredin23}, with an extension of the Whisper ASR model that performs diarization-informed target-speaker ASR~\cite{polok2025dicow}.
For multi-channel, we use the \emph{NTT CHiME-8}
system~\cite{kamo2025ntt_chime8} in its \texttt{small} (S) configuration, which combines channel
selection, guided source
separation~\cite{boeddecker18_chime}, EEND-VC with
multi-channel speaker counting, and target speaker voice activity detection (TS-VAD) refinement~\cite{TS-VAD}, followed by single-speaker ASR
with Whisper~\cite{whisper}.
Among the multiple systems~\cite{mitrofanov2024stcon,kamo24_chime,chime8,cornell2025recent} submitted to the CHiME-8 DASR challenge~\cite{cornell2025recent}, we select NTT~(S) specifically because it uses a single Whisper model without ensembling, making it more comparable to the DiCoW single-channel one. 
Although these systems perform slightly lower than the SOTA on CHiME-8 DASR tasks~\cite{MITROFANOV2025101780,mitrofanov2024stcon}, they both reflect the dominant designs in recent challenges~\cite{cornell2025recent, abramovski2025summary}.
\item \textbf{Task-Specific LLMs:}
We use VibeVoice~\cite{peng2026vibevoice} and Voxtral Mini Transcribe v2 (Voxtral MTv2)~\cite{voxtral2026} . Both systems fine-tune an LLM backbone for long-form multi-speaker transcription with speaker attribution, sacrificing general-purpose capabilities (e.g., summarization, QA) in favor of long-form transcription with speaker attribution, timestamps and optional context biasing.
Other LLM-based systems~\cite{yin2025speakerlm,huo2026tagspeech,yu2026moss} were excluded as they are limited to short segments~\cite{huo2026tagspeech,yin2025speakerlm} or are not publicly accessible~\cite{yin2025speakerlm, yu2026moss}.
\item \textbf{General-Purpose Multimodal LLM:}
Gemini~3.0 Flash~\cite{gemini3flash2025} is prompted to produce speaker-labeled, timestamped transcripts from single-channel
audio without task-specific fine-tuning.
Among this category, Gemini is currently the only one that natively supports long-form multi-talker transcription with segmentation and speaker attribution. Other models, such as Qwen3-Omni~\cite{xu2025qwen3}, lack built-in long-form multi-speaker support at the time of writing. 
\end{itemize}

\subsection{Datasets}



We use three benchmark datasets taken from the recent CHiME-8 DASR challenge~\cite{chime8}\footnote{Compared to~\cite{chime8}, we exclude
CHiME-6~\cite{watanabe2020chime} as its multi-room setup is prohibitively difficult for single-channel systems, and its sessions of more than two hours exceed the context length of most current E2E LLM-based systems~\cite{peng2026vibevoice, yu2026moss}.}, spanning diverse speaker counts (2--7), overlap ratios (12--30\%), acoustic conditions, and microphone configurations, covering three scenarios: interviews, office meetings, and informal dinner-party conversations.\\
\textbf{Mixer-6 (MX6)}~\cite{brandschain2010mixer}: two-speaker interviews ($\sim$14\% overlap, $\sim$25\,min avg.) recorded with 14 distributed microphones; single-channel uses CH04, multi-channel uses all non-close-talk channels (CH04--CH14).\\
\textbf{NOTSOFAR-1 (NSF1)}~\cite{vinnikov24_interspeech}: three-to-seven (eval set) speaker office meetings ($\sim$29\% overlap, $\sim$10\, min avg.) recorded on a single 7-channel array; single-channel uses \texttt{U01.CH1}.\\
\textbf{DiPCo}~\cite{van2020dipco}: four-speaker dinner parties ($\sim$25\% overlap, $\sim$35\,min avg.) recorded with five distributed 7-microphone arrays; single-channel uses \texttt{U01.CH1}. This is the most acoustically challenging scenario, with background music and reverberation due to a bigger room.


It is also worth noting that these datasets are less susceptible to transcript-level data leakage in LLM-based ASR systems, as recently documented for read-speech benchmarks~\cite{tseng2025evaluation}.
Our evaluation data are either behind restricted-access licenses or require non-trivial processing to extract, making incidental inclusion in web-crawled pre-training corpora unlikely.

\section{Evaluation Metrics}
\label{sec.metrics}


\subsection{Overlap-Aware cpWER and tcpWER}
\label{ssec:tcpwer}
\label{ssec:overlap_wer}

We consider CASR systems that output transcriptions with speaker, start time, and end time labels for each utterance.
cpWER~\cite{watanabe2020chime} concatenates all utterances per speaker, finds the speaker permutation minimizing the total Levenshtein distance, and normalizes by the total reference word count.
tcpWER~\cite{von2023meeteval, vonneumann2025wer} refines this by imposing a time constraint, such that reference--hypothesis word pairs separated by more than a collar $c$ (here, $c=5$\,s) are treated as errors in the Levenshtein distance computation, penalizing temporal misalignment.
Both metrics also yield the optimal speaker assignment and utterance-level alignment used by our extensions below.

Each reference utterance is annotated with start and end times, allowing us to determine whether it occurs in an overlapping region. Using the word-level alignments obtained from tcpWER, we can decompose the total number of word errors into those occurring in overlapping ($E^{\mathrm{ov}}$) and non-overlapping single-speaker ($E^{\mathrm{1spk}}$) segments.

Let $N_{\mathrm{ref}}$ denote the total number of reference words. Then,
\begin{equation}
\mathrm{tcpWER} =
\frac{E^{\mathrm{ov}} + E^{\mathrm{1spk}}}{N_{\mathrm{ref}}}, 
\qquad
\mathrm{tcpWER}_{\nu} =
\frac{E^{\nu}}{N_{\mathrm{ref}}},
\end{equation}
where $\nu \in \{\mathrm{ov}, \mathrm{1spk}\}$.

To assess recognition performance within each region independently of its relative frequency, we additionally define tcpWER normalized by each specific word count, $
\mathrm{tcpWER}_{\nu}^{\rm norm} =
E^{\nu}/N_{\mathrm{ref}}^{\nu},$
where $N_{\mathrm{ref}}^{\nu}$ denotes the number of reference words in overlapping ($\nu=\mathrm{ov}$) or non-overlapping ($\nu=\mathrm{1spk}$) segments.

$\mathrm{tcpWER}_{\nu}$ reflects the contribution of each region to the overall error, whereas $\mathrm{tcpWER}_{\nu}^{\rm norm}$ isolates the intrinsic recognition difficulty within that region.

\subsection{tcpSemER: Semantic Error Rate for Long-Form Multi-Talker Audio}
\label{ssec:tcpsemerr}

tcpWER penalizes all token-level deviations equally, regardless of semantic impact.
To assess whether errors alter meaning, we replace Levenshtein distance with a semantic similarity measure.
From the word-level alignment produced by tcpWER (Section~\ref{ssec:tcpwer}), we derive a set $\mathcal{A}$ of utterance-level pairs $(R,H)$, where each reference utterance $R$ is paired with the corresponding hypothesis segment $H$ (or with $\emptyset$ if $R$ is entirely deleted, and vice versa for inserted hypothesis segments).
For each pair, we compute a cosine similarity $\mathrm{SentSim}(R,H)$ between sentence embeddings from MiniLM-L12v2~\cite{wang2020minilm}%
\footnote{\href{https://huggingface.co/sentence-transformers/all-MiniLM-L12-v2}{huggingface.co/sentence-transformers/all-MiniLM-L12-v2, chosen for its favorable trade-off between speed and performance on the Massive Text Embedding Benchmark~\cite{muennighoff2023mteb}.}}.
The semantic error for a pair is defined as
\begin{align}
\mathrm{SemErr}(R, H) &\! =
\begin{cases}
|H|, & R = \emptyset, \\
|R|, & H = \emptyset, \\
(1\!- \mathrm{SentSim}(R, H))\cdot|R|,
  & \text{else},
\end{cases} \nonumber
\end{align}
where $|R|$ and $|H|$ denote word counts, and the semantic error rate is:
\begin{equation}
\mathrm{tcpSemER} =
\frac{\sum_{(R,H)\in\mathcal{A}}
  \mathrm{SemErr}(R,H)}{N_{\mathrm{ref}}}.
\label{eq:tcpSemER}
\end{equation}
This formulation preserves the permutation-invariant, time-constrained evaluation of tcpWER, but weights errors by semantic impact: meaning-altering deviations (e.g., negation deletion, entity substitution) produce low similarity and thus larger penalties, whereas surface-level variations (e.g., paraphrasing) incur smaller penalties.
Unlike tcpWER, tcpSemER is largely unaffected by surface-level transcription differences such as the inclusion or omission of fillers (``uh'', ``um''), stutters, and disfluencies, which are common in spontaneous speech and can vary significantly across systems and text normalization schemes. While such speech events are important for turn-taking prediction, social signal analysis~\cite{gravano2011turn}, or spoken dialogue model training and evaluation~\cite{aroralandscape}, they are less relevant when transcripts are consumed by downstream LLMs, e.g., summarization, QA, or spoken intent understanding~\cite{kim2021semantic}.
tcpSemER can thus complement tcpWER, not replace it: together they capture both verbatim and semantic
transcription accuracy.
Table~\ref{table:examples} illustrates this at the utterance level: errors that fundamentally alter meaning (e.g., negation deletion, entity substitution) incur higher penalties under tcpSemER than tcpWER ($\Delta < 0$), while surface-level variations such as filler insertion or paraphrasing are penalized more by tcpWER ($\Delta > 0$). 

\begin{figure}[t]
    \centering
    \includegraphics[width=0.9\linewidth]{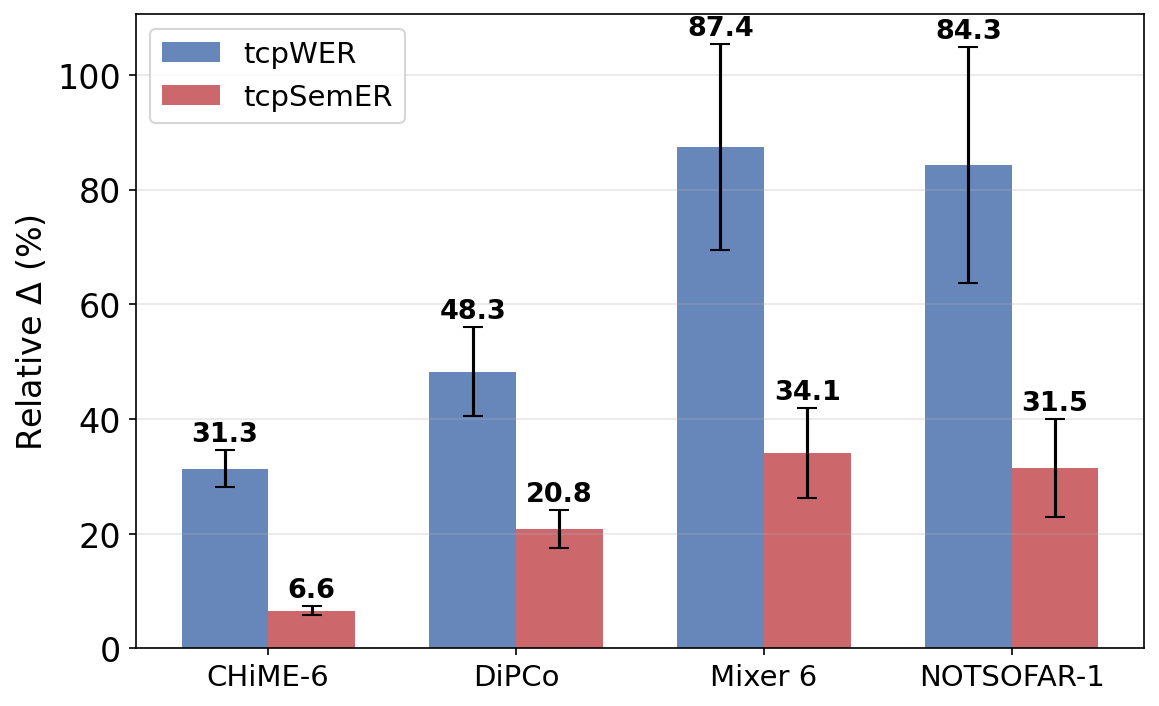}
   \vspace{-0.3cm}
    \caption{Relative change in tcpWER and tcpSemER when switching from CHiME-8 to CHiME-7 text normalization, averaged across the top-4 CHiME-8 DASR challenge systems per dataset. Error bars denote 
  standard deviation.}
    \label{fig:qa_tcpsemer}
\vspace{-0.3cm}
\end{figure}

Figure ~\ref{fig:qa_tcpsemer} plots the relative increases in error for tcpWER and tcpSemER when switching from CHiME-8 text normalization (more ``forgiving'', removes all non-lexical verbal sounds) to the CHiME-7 one (more verbatim). This relative error increase is averaged across CHiME-8 DASR challenge top-4 systems for each dataset (including CHiME-6). tcpWER is nore sensistive 
to the normalization scheme (17–56\% relative change) compared to tcpSemER (3–21\%), demonstrating that tcpSemER is
  more robust to surface-level text normalization choices though not entirely immune, since it inherits the tcpWER alignment.

\section{Experimental Analysis}
\label{sec:experiments}


\begin{table}[tb]
\centering
\caption{Representative utterance-level divergences between WER and SemER from top-3 NSF1 challenge systems. $\Delta = \mathrm{WER} - \mathrm{SemER}$; $\Delta > 0$: WER penalizes more; $\Delta < 0$: SemER penalizes more. \textcolor{red}{Red}: substitutions; \textcolor{blue}{blue}: insertions.}
\vspace{-3pt}
\label{table:examples}
\resizebox{\linewidth}{!}{
    \setlength{\tabcolsep}{4pt}
\small
\begin{tabular}{l p{2cm} p{2cm} c c c}
\toprule
\textbf{Category} & \textbf{Ref} & \textbf{Hyp} & \textbf{WER} & \textbf{SemER} & \textbf{$\Delta$} \\
\midrule
Negation & it is \textcolor{red}{lovely} & it is \textcolor{red}{not} & 0.33 & 0.67 & -0.34 \\ \hline
Entity & yeah that is a \textcolor{red}{mat} & yeah that is a \textcolor{red}{lot} & 0.20 & 0.83 & -0.63 \\ \hline
Filler & that is a great idea chris & that is a great idea chris \textcolor{blue}{yeah yeah yeah} & 0.50 & 0.05 & 0.45 \\ \hline
Paraphrase & \textcolor{red}{or maybe} like a slogan & \textcolor{red}{that could} \textcolor{blue}{be} like a slogan & 0.60 & 0.12 & 0.48 \\
\bottomrule
\end{tabular}}
\vspace{-0.6cm}
\end{table}

\subsection{Overall Results}
\label{ssec:overall}
\autoref{table:wer_der} summarizes cpWER, tcpWER, tcpSemER, and the diarization error rate (DER) across all datasets.
For the single-channel cases, the specialized LLM model VibeVoice achieves reasonable performance on MX6 and NSF1, although it is about 10--40\% worse than DiCoW in terms of cpWER and tcpWER (rows 1 vs.\ 2, and 8 vs.\ 9). Interestingly, for MX6, the DER of VibeVoice is better than that of DiCoW (rows 1 and 2). However, VibeVoice performs much more poorly on DiPCo (row 16). Voxtral MTv2 (rows 3, 10) could perform diarization to some extent for MX6 and NSF1, but failed on several DiPCo sessions ($>50$ speakers recognized), preventing us from scoring it. The general-purpose LLM model, Gemini~3.0, performs significantly worse overall. 
Gemini 3.0 approaches VibeVoice only on NSF1 cpWER (row 11) but shows much higher tcpWER across all datasets, reflecting poor diarization.

DiCoW and VibeVoice do not support multi-channel inputs. To assess the impact of multi-channel processing, we transcribe each microphone independently and apply Meeting recognizer Output Voting Error Reduction (MOVER)~\cite{kamo25_interspeech} to combine the multi-speaker transcriptions.
Combining channel-wise results with MOVER consistently improves performance for both VibeVoice and DiCoW (rows 6, 7, 13, 14, 20, and 21). Notably, on MX6, these systems (rows 6 and 7) outperform the top challenge system, which achieved a tcpWER of 10.9\% \cite{mitrofanov2024stcon}. However, modular multi-channel systems with an array processing front end, such as the CH8 DASR NTT system (rows 5, 12, 19), remain clearly stronger when conversational complexity increases (more speakers, higher overlap, more adverse acoustics).

The SemER columns generally follow the trends observed in WER, but with some differences. On MX6, VibeVoice (row 2) has a higher tcpWER but a lower tcpSemER than CH8 DASR NTT (row 5). This suggests that a larger share of VibeVoice’s errors are surface-level variations (e.g., fillers or disfluencies) rather than meaning-altering deviations.

These results show that specialized LLMs are becoming competitive alternatives to modular approaches for two-speaker conversations, but modular systems remain significantly better for more complex conversations. The positive results with simple multi-channel integration suggest the need for more research into multi-channel LLM systems for CASR.

\begin{table}[tb]
\caption{Results on all datasets. Collars are 5\,s (tcpWER) and 0.25\,s (DER). All values are in \%. For each dataset, we report the number of speakers, the overlap ratio, the microphone setup, and the average meeting duration.}
\vspace{-0.2cm}
   \sisetup{
    round-mode = places,
    round-precision = 1,
    table-format = 3.1
}
\label{table:wer_der}
\resizebox{0.99\linewidth}{!}{
\setlength{\tabcolsep}{4pt}
\small
\begin{tabular}{@{}l@{\hspace{0.08cm}}l cc S@{\hspace{0.08cm}}S SS @{}}
\toprule
& & & & \multicolumn{2}{c}{WER} & {SemER} & {DER} \\
 \cmidrule(lr){5-6}
&System & LLM &Ch. & {cp}&{tcp}&{tcp}&\\
\midrule
\rowcolor{gray!20}\multicolumn{8}{c}{\textbf{MX6} {\small (2\,spk, 14\%\,ovl, distributed, $\sim$15\,min)}} \\
1 &DiCoW& --&1ch&14.48&14.71&7.84&21.05\\
2 &VibeVoice    & \checkmark & 1ch & 16.03 & 16.32  & 7.51&9.89\\
3 &Voxtral\,MTv2 &\checkmark &1ch&  19.37 & 19.59&10.46& 21.97\\
4 &Gemini 3.0   & \checkmark & 1ch & 58.34 & 115.54 &80.79 & 78.49\\
\midrule
5 &CH8 DASR NTT (S)& --&mch&14.96&15.32&8.06&8.26\\
6 &DiCoW + MOVER & --& mch & 9.92 & 10.11 & 4.90& 21.26\\
7 &VibeVoice + MOVER & \checkmark& mch &10.62&10.79&4.68&13.10\\
\midrule
\rowcolor{gray!20}\multicolumn{8}{c}{\textbf{NSF1} {\small (3--7\,spk, 29\%\,ovl, single array, $\sim$6\,min)}} \\
8 &DiCoW & -- &1ch&23.18&24.55&15.70&7.65\\
9 &VibeVoice& \checkmark&1ch&35.75&36.62&22.47&18.5\\
10 &Voxtral\,MTv2 &\checkmark &1ch& 54.39 &58.14 &38.67 &29.95\\
11 &Gemini 3.0& \checkmark&1ch&39.14&126.28&75.58&74.71\\
\midrule
12 &CH8 DASR NTT (S)& --& mch & 14.47 &  15.03 &  9.46 & 10.91\\
13 &DiCoW + MOVER& --&mch& 20.36 & 21.21 & 14.35& 9.44\\
14 &VibeVoice+MOVER& \checkmark&mch&29.10&29.47&18.03& 18.57\\
\midrule
\rowcolor{gray!20}\multicolumn{8}{c}{\textbf{DiPCo}
{\small (4\,spk, 25\%\,ovl, distributed, $\sim$31\,min)}} \\
15 &DiCoW& --&1ch&34.03&36.83&23.25&27.15\\
16 &VibeVoice& \checkmark&1ch&68.80&70.72&42.53&35.81\\
17 &Voxtral\,MTv2 &\checkmark &1ch&  \multicolumn{4}{c}{---\emph{failed}---}\\
18 &Gemini 3.0& \checkmark& 1ch & 91.22 & 111.14 & 97.26 & 98.44\\
\midrule
19 &CH8 DASR NTT (S)& --&mch&23.64&25.02&13.98&20.33\\
20 &DiCoW + MOVER& --&mch&  29.72& 31.15& 22.04&34.06 \\
21 &VibeVoice + MOVER& \checkmark&mch&58.37&59.17&35.90&41.00\\
\bottomrule
\end{tabular}}
\vspace{-0.5cm}
\end{table}

\subsection{Effect of Overlap}
\label{ssec:overlap}

Table~\ref{table:overlap} decomposes tcpWER into contributions from overlapping and single speaker regions for the most representative systems.
Across all datasets, the majority of errors originate from overlapping segments.
For example, on NSF1, although 32\% of segments contain overlap (after tcpWER alignment), these segments account for about 90\% of the total error (rows 5--8), indicating that overlap is the dominant source of difficulty. 

Looking at the normalized $\mathrm{tcpWER}^{\rm norm}$ metrics (${\mathrm{ov}}^{\rm norm}$ and ${\mathrm{1spk}}^{\rm norm}$ columns), we observe that, in most cases, performance is worse in overlap segments than in single-speaker segments. However, in some cases, e.g., CH8 DASR NTT system for MX6 (row 3), recognition performance on single speaker segments is worse, probably due to speaker counting errors. VibeVoice performs significantly worse than modular systems in overlap segments. This illustrates the difficulty of resolving speaker overlap within its LLM-based architecture without explicit diarization or separation components.

These results highlight that overlap handling, rather than single-speaker recognition accuracy, remains the principal bottleneck in conversational ASR.

\begin{table}[tb]
\caption{Overlap-aware tcpWER decomposition (\%). $\mathrm{ov}$/$\mathrm{1spk}$: contribution to total error; $\mathrm{ov}^{\rm norm}$/$\mathrm{1spk}^{\rm norm}$: normalized by region-specific word count.}
   \sisetup{
    round-mode = places,
    round-precision = 1,
    table-format = 3.1
}
\vspace{-0.3cm}
\label{table:overlap}
\resizebox{\linewidth}{!}{
\setlength{\tabcolsep}{4pt}
\small
\begin{tabular}{@{}l@{\hspace{0.08cm}} l c S S@{\hspace{0.08cm}}S S@{\hspace{0.08cm}}S @{}}
\toprule
& & & \multicolumn{5}{c}{tcpWER (\%)} \\
 \cmidrule(lr){4-8}
&System &Ch. & {all}&{ov}&{1spk}&{ov$^{\rm norm}$}&{1spk$^{\rm norm}$}\\
\midrule
\rowcolor{gray!20}\multicolumn{8}{c}{\textbf{MX6}
{\small ($N_{\mathrm{ref}}^{\mathrm{ov}}=37628$, $N_{\mathrm{ref}}^{\mathrm{1spk}}=32990$) }} \\
1& DiCoW &1ch&14.71 & 8.50 & 6.21 & 15.96 & 13.29\\
2&VibeVoice&1ch&16.32 & 10.73 & 5.59 & 20.13 & 11.97\\
\midrule
3&CH8 DASR NTT (S)       &mch &15.32 & 7.56 & 7.76 & 14.19 & 16.61 \\
4&VibeVoice + MOVER&mch&10.79 & 7.05 & 3.74 & 13.22 & 8.01 \\
\midrule
\rowcolor{gray!20}\multicolumn{8}{c}{\textbf{NSF1}
{\small ($N_{\mathrm{ref}}^{\mathrm{ov}}=176608$, $N_{\mathrm{ref}}^{\mathrm{1spk}}=59132$)}} \\
5&DiCoW &1ch&24.55 & 22.91 & 1.64 & 29.01 & 7.81\\
6&VibeVoice&1ch&36.62&33.22&3.4&44.34&13.54\\
\midrule
7&CH8 DASR NTT (S)       & mch & 15.03 & 13.42 & 1.61 & 17.91 & 6.42 \\
8&VibeVoice + MOVER&mch&29.47 & 26.63 & 2.84 & 35.55 & 11.33 \\
\midrule
\rowcolor{gray!20}\multicolumn{8}{c}{\textbf{DiPCo}
{\small ($N_{\mathrm{ref}}^{\mathrm{ov}}=20592$, $N_{\mathrm{ref}}^{\mathrm{1spk}}=8238$) }} \\
9&DiCoW&1ch&36.83 & 26.94 & 9.90 & 37.71 & 34.63\\
10&VibeVoice&1ch&70.72 & 56.01 & 14.71 & 78.42 & 51.48\\
\midrule
11&CH8 DASR NTT (S)&mch        & 25.02 & 18.75 & 6.26 & 26.27 & 21.90\\
12&VibeVoice + MOVER&mch&59.17 & 41.62 & 17.56 & 58.27 & 61.45\\
\bottomrule
\end{tabular}}
\vspace{-0.4cm}
\end{table}

\begin{figure}[t]
    \centering
    \includegraphics[width=.99\linewidth]{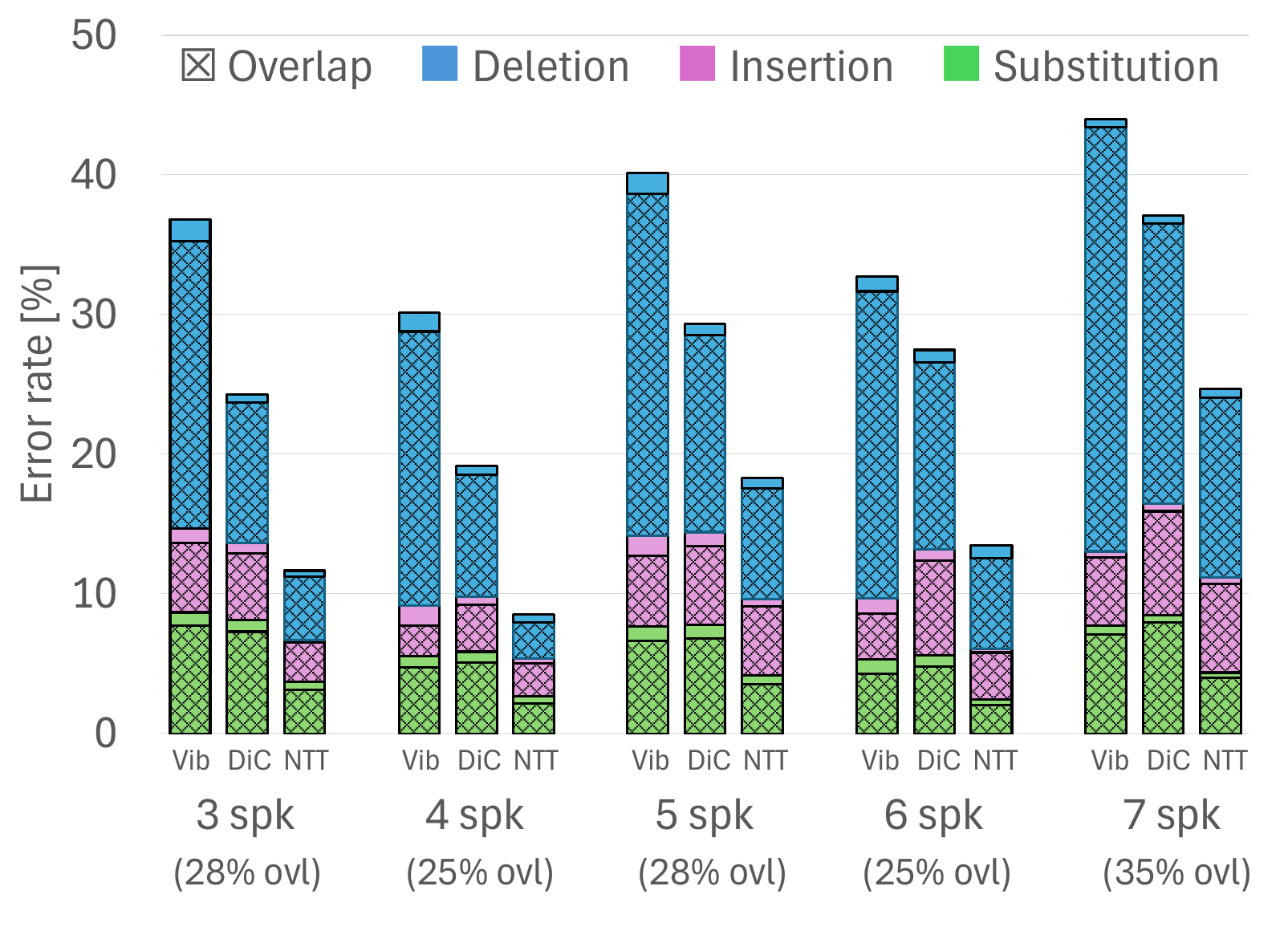}
    \vspace{-0.2cm}
    \caption{Decomposition of tcpWER into deletion, insertion, and substitution errors for VibeVoice (Vib), DiCoW (DiC), and CH8 DASR NTT (NTT) on NSF1, shown by speaker count. The proportion of overlapped speech (ovl) is shown in parentheses. }
    \label{fig:tcpWER_NSF1}
\vspace{-0.4cm}
\end{figure}

\subsection{Impact of Number of Speakers and Overlap Ratio}

Figure~\ref{fig:tcpWER_NSF1} illustrates the decomposition of tcpWER into deletion, insertion, and substitution errors for DiCoW, VibeVoice, and CH8 DASR NTT (S) on NSF1 as a function of speaker count, along with the corresponding overlap ratios.
Because higher speaker counts are typically associated with higher overlap ratios in NSF1, these two factors are analyzed jointly in this figure.
Across all systems, the overall error rate increases as the number of speakers and overlap ratio grow, indicating that speaker overlap remains a major source of recognition errors.

VibeVoice exhibits a particularly high number of deletions in overlap regions.
Qualitative inspection revealed that, during overlapping speech, it often recognizes only a single dominant speaker while omitting concurrent speakers, especially short, fully overlapped backchannels.
In contrast, DiCoW shows a more gradual increase in deletions as speaker count and overlap increase, indicating greater robustness to speaker interference.

When the number of speakers exceeds four, DiCoW shows a larger increase in insertion errors than VibeVoice.
These errors may be related to incorrect speaker estimation.
Indeed, VibeVoice achieves better speaker-counting accuracy than DiCoW (77.5\% vs. 71.9\%) and lower MAE (0.269 vs. 0.319).
In contrast, DiCoW tends to overestimate the number of speakers, introducing redundant speaker streams that are counted as insertions under tcpWER.

The NTT system achieves the highest speaker-counting
accuracy (85.6\%, MAE 0.15) by leveraging multi-channel
inputs, and consistently exhibits the lowest error rates across all categories, confirming that modular systems employing explicit separation before single-speaker ASR remain the most robust strategy for complex
conversational scenarios.

\section{Concluding Remarks}
\label{sec:conclusion}
We systematically compared modular pipelines, task‑specific LLMs, and general‑purpose multimodal LLMs for CASR across three datasets of increasing complexity, using newly introduced evaluation metrics. Our analysis confirms that overlap handling remains the primary challenge in CASR. Modular systems, which include explicit diarization and separation front‑ends, outperform LLM‑based systems mainly because they handle overlap more effectively.
Task‑specific LLM‑based systems perform competitively in two‑speaker scenarios, particularly in terms of tcpSemER, suggesting that their errors tend to be surface‑level rather than meaning‑altering. However, their performance degrades more sharply than modular systems as overlap, speaker count, and acoustic difficulty increase. Simple multi‑channel integration via MOVER helps narrow the gap between LLM-based and modular systems, motivating further research into native multi‑channel LLM‑based models for CASR.
Finally, general‑purpose multimodal LLMs do not yet appear ready for CASR, especially due to insufficient diarization capabilities.

\section{Generative AI Use Disclosure}
This manuscript was edited and polished with the assistance of generative AI.
Generative AI models were also used as comparison systems in the experimental evaluation.
All experimental design, implementation, and analysis were conducted by the authors, who take full responsibility for the content.
\bibliographystyle{IEEEtran}
\bibliography{mybib}
\end{document}